\documentclass[twoside,11pt]{article}
\usepackage[letterpaper]{geometry}
\usepackage{jmlr2e,graphicx,times,amsmath} 

\ShortHeadings{Sparse, guided feature connections}{Knittel and Blair}
\firstpageno{1}

\def \AND {`and' }
\def \OR  {`or' }

\begin{document}

\title{Sparse, guided feature connections in an Abstract Deep Network}

\author{\name Anthony Knittel \email a.knittel@unsw.edu.au \\
       \addr University of New South Wales
       \AND
       \name Alan Blair \email blair@cse.unsw.edu.au \\
       \addr University of New South Wales}

\editor{}

\maketitle

\begin{abstract}
We present a technique for developing a network of re-used features, where the topology is formed using 
a coarse learning method, that allows gradient-descent fine tuning, known 
as an Abstract Deep Network (ADN).
New features are built based on observed co-occurrences, and the network
is maintained using a selection process related to evolutionary algorithms.
This allows coarse exploration of the problem space, effective for 
irregular domains, while gradient descent allows precise solutions.
Accuracy on standard UCI and Protein-Structure Prediction problems is comparable with benchmark SVM and optimized
GBML approaches, and shows scalability for addressing large problems.
The discrete implementation is symbolic, allowing interpretability, while the continuous method
using fine-tuning shows improved accuracy.
The binary multiplexer problem is explored, as an irregular domain that does not support gradient
descent learning, showing solution to the benchmark 135-bit problem.
A convolutional implementation is demonstrated on image classification, showing an error-rate of 
0.79\% on the MNIST problem, without a pre-defined topology. 
The ADN system provides a method for developing a very sparse, deep feature topology, based on observed
relationships between features, that is able to find solutions in irregular domains, and initialize 
a network prior to gradient descent learning.
\end{abstract}

\section{Introduction}

A goal of an artificial learner is to discover a model of its environment,
that captures essential properties for its goals, and allows prediction
of unseen attributes.
Although shallow techniques can provide arbitrary accuracy with enough training examples and resources~\citep{hecht1989theory}, the use of a deeper model, where intermediate features are found, allows structure in the environment to be captured, with advantages for efficiency and re-use~\citep{bengio2009learning}.  
Representations in the brain are commonly seen as having re-used elements in a deeper structure, for example in the visual cortex, where neurons responding to simple features are combined into increasingly complex structures~\citep{hubel1968receptive}.  Abstract semantic concepts share similar properties, where activation of one concept can facilitate another, as a result of shared structures and connections~\citep{mcnamara2005semantic}. The use of re-used elements in the brain, 
is a demonstration of the effectiveness of this model for representing a large and diverse collection of features and objects, that can be recognized quickly in a wide range of environments.  

Ideally, artificial learning should be able to capture information from a wide range of sources and modalities, identify common structure, both unsupervised and in a task-relevant way, and to integrate such information, including the contextual relationships between features.  Traditional neural network approaches focus on highly localized mechanisms, such as gradient descent from reconstruction of a single layer, or from back-propagation of the output of a handful of classification units.  These processes are important, however processes can be recognized in the brain that act on a larger scale, such as patterns observed in behavioural experiments, that give clues about broader-scale processes acting on larger populations of neurons.  Examples include properties of how concepts are reinforced and accessed~\citep{anderson04}, how activation can pass between related concepts~\citep{mcnamara2005semantic}, and the manner in which context can influence the perception of parts~\citep{bar04}.  
The presence of re-used structures (the broader topic addressed in deep learning) can also be seen in behavioural studies, such as in patterns or chunks used by expert players of Go or Chess, where task-relevance of features is essential~\citep{chase1973perception, didierjean08}.  
An emphasis on including features that are significant for discrimination is also addressed in studies of decision making~\citep{gabaix}. 
These topics are typically studied independently, and offer very diverse perspectives, however address aspects of cognition with many common elements.  A broader aim of the topic this study is exploring, is to identify common traits emerging from these diverse experiments, to aid in the design of artificial learning systems.  




In this paper we present a novel learning system called the Abstract Deep Network (ADN), that develops a deep, re-used structure of features useful for artificial learning\footnote{Code is available at \texttt{github.com/anthonyk91/ADN}}.  This is based on processes functionally related to evolutionary algorithms, such as selection acting on a `population' of features, representative of processes acting on neurons at a broader scale.  
The internal connections (weights) of the ADN can either be discrete (fixed)
or continuous, in which case they may be tuned by gradient descent.
We show how the ADN system can construct a feature network
able to find solutions in regular and irregular search spaces, 
developing features appropriate for the task at hand.



The paper is organized as follows: Deep learning neural networks and Learning Classifier Systems
are reviewed in Sections 2 and 3. The Abstract Deep Network is
described in its discrete and continuous form in Sections \ref{adndesign} and \ref{gradientdescent}.
The ADN system is then demonstrated on a wide variety of tasks
including UCI classification and Protein Structure Prediction
(Section~\ref{pointfeatures}), irregular binary multiplexer classification
(Section~\ref{irregproblem})
and image processing (Section~\ref{secimages}) followed by concluding remarks in
Section~\ref{secconc}.

\section{Deep learning neural networks}

Deep neural networks are designed with the aim of capturing re-usable features, in a structure that allows re-combination of discovered elements.  One important method is the use of layer-wise pre-training, where a layer of simple features is discovered using unsupervised learning from the observed data, and subsequent layers capture more complex structures, based on the statistical relationships of features in the lower level~\citep{erhan2010does}.  
For a given observation each subsequent layer refines the activity of a previous layer, using a learned set of statistics to encourage or discourage patterns of features according to prior observations, when the network is viewed from the perspective of attempting to reconstruct an observed input.  The features that are discovered are recurring structures in the input data, that are useful for allowing an observation to be reconstructed, independent of other tasks and objectives of the system.

Another common and successful approach in deep neural networks is convolutional networks~\citep{lecun98}, that are constructed with a fixed topology, commonly using layers of tuning, pooling, contrast normalization and rectification functions.  The topology borrows from properties of cortical structures, and can be very effective for identifying visual features with a degree of scale, translation and lighting invariance.  The topology used is a key aspect of their success, as the use of a specific topology, even with randomly constructed weights, can achieve near-benchmark performance~\citep{jarrett2009best}.  Visual processing can be considered a specialized domain, and dedicated networks are present in mammalian brains, however such approaches do not necessarily provide a general learning technique, that allows capturing structure and relating common features in a more general way.




\section{Learning Classifier Systems}



Learning Classifier Systems are a family of evolutionary methods, based on a population of classifiers that each relate sections of the feature space with a classification and a measure of accuracy, and typically explore new solutions using Genetic Algorithms~\citep{holland2000learning}.  In Michigan-style systems, each classifier responds to a region of the feature space, which may be large or small, and for each observation multiple classifiers will offer predictions, with varying degrees of accuracy.  



In this way the population describes a number of generalized and specialized interpretations of the feature space.  A general rule offers a prediction for a large space with possibly limited accuracy, while specialized rules may offer more accurate predictions for specific regions, that can refine more general predictions.  The relationship between classifiers is not necessarily based on subsumption, and varying degrees of overlap may occur.  Appropriate integration of the predictions of different rules matching a given observation is non-trivial, and is typically addressed as a weighted average.  The use of general and specialized predictions is loosely related to `levels of processing' in cognitive studies, where an object such as a robin can be interpreted at the basic level as a `bird', or at a more specific level as a `robin', with the identification at different levels occurring in parallel~\citep{scott2006reevaluation}.  In Learning Classifier Systems multiple interpretations may be active at once, with each responding to a set of observed features independently.  



LCS, like other evolutionary systems, are effective at finding a good solution in highly irregular environments.  They have shown success on problems such as Protein Structure Prediction~\citep{Bacardit2009}, robotic control~\citep{butz2008context}, medical diagnosis~\citep{llora2007towards}, and on artificial tasks where gradient descent does not allow discovery of the solution~\citep{iqbal2012reusing}.  
The classification features that are discovered are task-focused, based on the ability to discriminate between classes and improve the overall accuracy of the system.  As a further advantage LCS systems provide interpretable solutions, however they are often not as able to identify precise solutions as gradient descent methods like artificial neural networks.


Important features of the environment are captured in classifiers, as a description of the region of interest the classifier responds to.  The Genetic Algorithm allows recombination, and preserves significant feature descriptions, which are stored redundantly in multiple classifiers in the system.  Significant features in the genetic code are known as building blocks~\citep{goldberg}, and it is preferable for these blocks to be preserved by the crossover mechanism, however they are not typically defined explicitly.  This places limitations on the ability to recombine significant features, as the encoding limits  independence and sparsity.


\section{Abstract Deep Network Design}
\label{adndesign}
\begin{figure}[ht]
\begin{center}
\centerline{\includegraphics{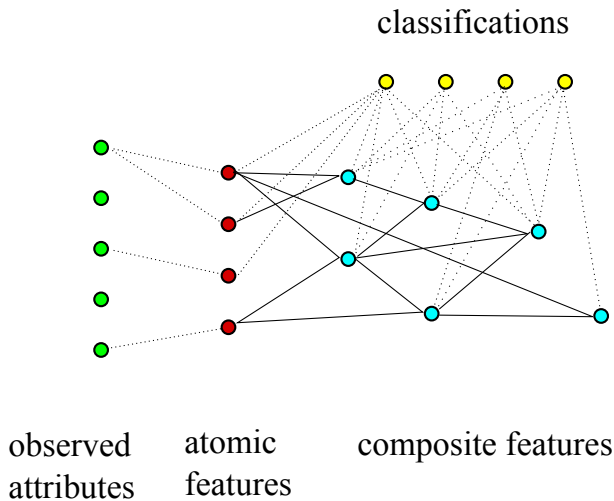}}
\caption{Structure of the feature network, showing atomic features responding to observed attributes, composite features based on child features, and relationships between features and classification nodes.}
\label{figoverview}
\end{center}
\end{figure}

The Abstract Deep Network (ADN) system we present shares a number of properties with Learning Classifier Systems, using a population based method to develop a hierarchical network of re-used features.
Each feature responds to a particular region of the input space, ranging from very general at the shallowest (atomic) level to highly specific at deeper levels.
All features are connected by continuous weights to a final output layer, used for classification.
In the Discrete version of ADN, the internal weights remain fixed and only the
weights in the final (output) layer can be tuned.
In the Continuous version (described in Section~\ref{gradientdescent})
the internal weights can also be tuned, by gradient descent,
in a manner comparable with artificial neural networks.
Figure~\ref{figoverview} provides an overview of the structure of the feature network.

\subsection{Feature population}

At the most basic level, an atomic ADN feature is an element that can be related directly with an observation.
For example, it might correspond to a certain
discrete (binary or categorical) input attribute having a specific value,
or it might indicate that a continuous input attribute satisfies a
certain inequality.

Composite (parent) features are constructed as combinations of existing (child) features, which may be either atomic or existing composite features.
In the current ADN implementation each parent node is a conjunction (logical `and')
of two constituent child nodes.
The final output of the system is a linear combination these conjunctive
features (and can therefore be seen as a kind of generalized Disjunctive Normal Form).

LCS classifiers are based on a relationship between a feature description, a classification (or action), and a measure of accuracy.  In contrast ADN features, which respond to certain aspects of the input space, are defined independently of classifications, and relationships are maintained between each feature and multiple (or no) classifications, as a result of network connections.

\begin{figure}[ht]
\begin{center}
\centerline{\includegraphics[width=\linewidth]{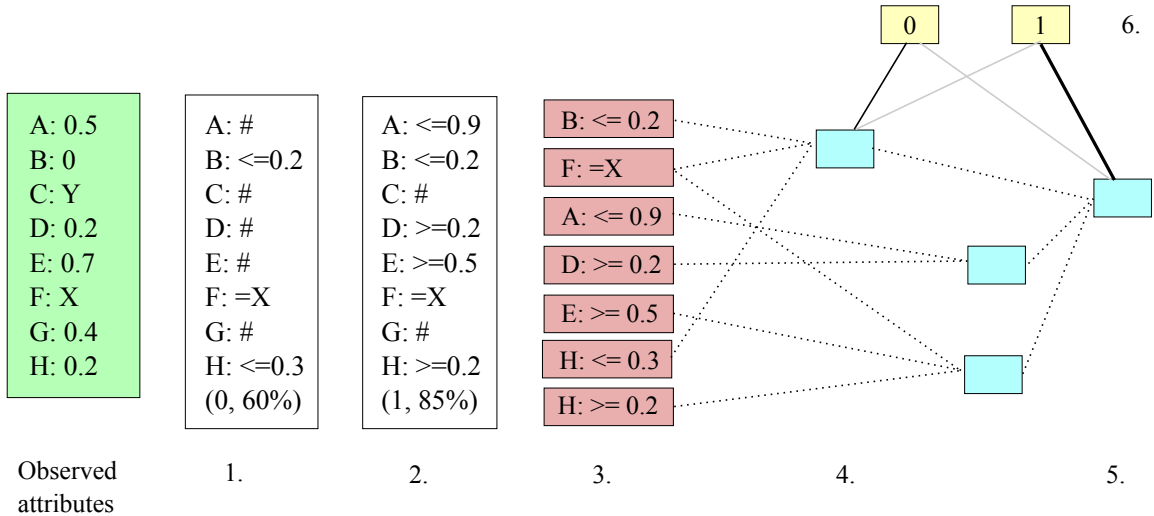}}
\caption{Relationship between a network of logical features and comparable LCS classifiers.  Of the observed attributes, C and F are nominal, and the rest are real-valued.  1. Matching LCS-style classifier (general), 2. Matching LCS-style classifier (specialized), 3. Set of matching ADN-style atomic features, 4. ADN composite feature equivalent to (1), 5. ADN composite feature equivalent to (2), 6. Output units, showing relationships with (4) and (5).}
\label{figlcsadn}
\end{center}
\end{figure}

Features can be interpreted as LCS classifiers, through the combination of child elements and connected classifications, and as such the structure can be considered as an alternative encoding of the recombination of various building blocks.  Figure~\ref{figlcsadn} describes the relationship between a number of LCS classifiers and an equivalent encoding based on re-used features in ADN.  An essential difference between the encoding used in ADN and building blocks in Genetic Algorithms, is that feature representations are independent, providing a sparse representation that allows explicit re-use and recombination of features.

The use of \AND operations for composite features results in a network of incrementally specialized features.  Low level features have few constraints, and respond to large regions of the space, acting as general classifiers.  Higher level features that combine a number of elements are more constrained in their response, providing specialized interpretation.  Through combination of different child elements, responses to various regions can be defined.  

Candidate \AND relationships can be identified from observed co-occurrences of features.  Meaningful \OR relationships are more difficult to identify through local operations, as individual instances are not indicative of significant disjunctions, and it is likely that retrospective analysis of some kind is necessary to identify meaningful \OR relationships.  

There may be benefits with the use of alternative operations, such as the use of \OR features to allow generalizations~\citep{si2013}, however for the purpose of this study a population based on \AND relationships is used.  This has been shown to have advantages in restricting the search of parent elements, and no advantage has been shown with the use of additional operators on the problems examined.  On logic problems, the use of a network of \AND operations can be considered related to Disjunctive Normal Form 
(depending on the response function developed by the classification layer).

Using this approach, low level features respond to large regions of the state space, while higher level features define increasingly specialized regions, based on the conjunction of lower level elements, 
that capture the intersection of the regions defined by each child element.  It is possible to arbitrarily divide the feature space using this approach. 
Whether this provides an efficient way of representing the feature space depends on the problem domain.  It is possible to construct problems where this representation would be inefficient, however in practise it has proved to be an effective method for feature description. 

\subsection{Classification relationships}


The relationship between features and classes is based on logistic or softmax regression, similar to the top layer commonly used in artificial neural networks, in contrast to the weighted average common to LCS systems such as XCS~\citep{xcsclassfit}.  The methods are related, each is used to balance the interpretations provided from a number of active features or classifiers.  In XCS each classifier defines an independent probability value for the accuracy of classification, while in an ADN the weight value between each feature and classification node is defined through gradient descent, and is interdependent on the weights of other co-active features.  For a two-class problem a single weighted edge is used between each feature and the output node, where the weight value represents the significance of an active feature for classification.  This mechanism allows accurate, significant features to supersede more general ones.

In LCS, each classifier relates a particular classification with a given region of the feature space.  
This can be considered a sparse representation between features and classes, as a number of regions of interest are defined, and each classifier relates a region with one classification.  
In the ADN system features are handled separately as a population, and a complete set of connections between features and classifications is used.  This allows the development of the feature population to take place independently, without an additional selection process for choosing particular relationships between features and classes.  The number of connections between features and classes has been found to be manageable in most cases, however an additional process to allow a sparse set of connections between features and classes is described in Section~\ref{secsparse}.

Weights between features and classes are initially set to zero, and are modified through gradient descent learning~\citep{haykin2009neural}.  From the activations of the classification nodes, and the target activations, weights are adjusted according to the derivative of the error function.  Bias values are used for each classification node.

It is not necessary to use random weights, the purpose of such weight initialization is typically to break symmetry, and as features are constructed with a semantic basis, difficulties with symmetry generally do not arise.  Use of initially zero weights is helpful in a dynamic population, as it allows new features to be introduced in a manner with initially zero interference with the existing network, and influence of the feature is introduced gradually through learning, based on the relative contribution of the feature compared with the existing population.

\subsection{Activation}

The use of conjunction relationships to define features, in a network of re-used elements of increasing specialization, is efficient for activating relevant features.  For an observation, examination takes place in a bottom-up pass, that only continues through features that have been activated, discontinuing search of all parents of an inactive feature.  One of the most expensive operations for LCS systems is to identify the classifiers that match the current observation, and this operation is considered one of the limiting factors of existing designs.  The design of the ADN feature network and method of activation 
addresses this challenge, as activation of a feature only needs to take place once, instead of being repeated for each redundant copy, and only a limited section of the network is examined.

\OR relationships have not been used, the introduction of such relationships greatly increases the number of active features for a given pass. 
If further complex operators are used, such as inverse relationships between features, a complete search of the feature network would be required.  When soft operations are used rather than logical relationships, for example when the system is operating as an artificial neural network as described in Section~\ref{gradientdescent}, a complete bottom-up pass of the network is necessary.

\subsection{Creation and selection}

The feature network is dynamic, new features are routinely added to the network, and weaker features are removed to bound the size of the population.  After an observation is made a number of existing features will be active.  For each observed instance, new features are created with a fixed probability, with a slight bias towards creating new features more frequently after failed classifications (about 75\% vs 25\% ratio).  New atomic features are created that match the input, for example the value of the $n^{th}$ attribute being greater or equal to the current value, and new composite features are constructed using a random set of active features as children.  In this way new features are constructed based on observed co-occurrences.  

When a new feature is added, a number of existing features are examined to test if an identical feature exists.  From the child features used to construct the new feature, only the existing next-level (parent\footnote{The term parent is used to refer to a higher level node in the tree connected to a lower child node, although the term can be misleading as the tree is constructed from bottom to top, and as such `children' are present before `parents'.}) features are examined to check for duplicates.

Many LCS systems use biased selection of classifiers, in order to construct new rules based on the most `fit' rules, encouraging the re-use of successful elements.  For simplicity, in the following experiments no weighting has been used in the selection of child features, each is chosen with equal probability.  From the set of chosen features, typically using only two child elements, a new composite element is constructed.  

To maintain a bounded size population, at a regular interval features are removed.  The value used for selection (equivalent to `fitness'), is based on the norm of the classification weights, taking into account the value of all parents of a given feature.  This measure reflects the significance of each feature for classification, such that features are removed that have the least influence on classification.  The selection measure is determined as follows:

\begin{equation}
f_n = \max(\left|W_n\right|, f_p),\qquad p \in \mathrm{parents}(n) 
\end{equation}

where $f_n$ is the selection value for feature $n$, and $W_n$ is the set of weights between feature $n$ and the classification nodes.  The L$_1$ norm of the classification weights has been used for simplicity.  

\section{Continuous Network with Gradient Descent}
\label{gradientdescent}

The feature network as described in Section~\ref{adndesign} follows a strictly logical design, where elements provide binary match responses at the atomic level, and subsequent evaluations are conducted in a discrete manner (until the classification layer, where real-valued interpretations are used).
The system can alternatively be viewed as a continuous (neural) network,
and the weights at all levels of the network can be tuned through gradient descent.

Two continuous system designs are presented.  The first is based on the point-feature design (Section~\ref{pointfeatures}), where data attributes are either real-valued or from nominal sets, and atomic features respond as $\leq$ or $\geq$ for real attributes, and $=$ for nominal attributes.  Composite features respond according to the conjunction of child activations.  The first gradient descent design provides a continuous implementation of the logical (discrete) design.  This allows a logical system to be implemented, which allows fast training and produces interpretable rules, and subsequently fine-tuned as a continuous system.  Alternatively the population learning technique can act in tandem with the gradient descent learning, providing influences on learning from different perspectives, where the population learning alters the topology based on coarse learning principles, while gradient descent provides fine-tuning adjustment of weights.  Note that as a continuous system, interpretability of rules is no longer possible, other than as a real-valued network.

The second continuous design is independent of the logical system.  Features are initialized using random weights, and are modified such that newly created features respond as matching the current observation.  Matching is defined according to a threshold, so each feature has a continuous activation value, and a discrete match value.  Composite features are defined using a soft-and function, and initialized such that the feature matches according to the activation of its child elements.

Both of these approaches are distinct from the method used by NEAT~\citep{stanley2002evolving} and other forms of evolutionary development of neural networks.  NEAT is an evolutionary approach for construction of a neural network topology using Genetic Algorithms, where each member in the population is a complete network, and refinement takes place by selection over multiple independent networks.  In contrast, our method is based on a population of features, where each is independently selected for usefulness according to local rules, and new features in the population are constructed based on combinations of existing features.

\subsection{Soft-logic gradient descent (first method)}

Capturing the logical design using a soft representation is performed using a soft-boundary response, where each atom is implemented using a sigmoid (tanh) activation function, and the parameters are chosen to approximately capture a definition such as `$\geq 3.2$'.  The soft margin is specified as a hyper-parameter, and should approximately reflect the expected margin between values of a given attribute (after normalization).  

\begin{figure}[ht]
\begin{center}
\centerline{\includegraphics[width=.7\columnwidth]{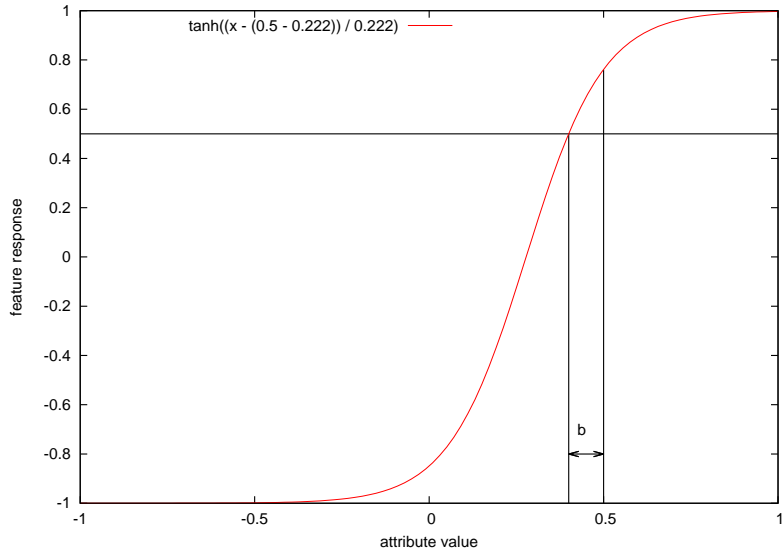}}
\caption{Method for constructing a soft response function related to $>=$ and $<=$ operators, acting on real-valued attributes.  The horizontal line represents the activation threshold, above which the feature is considered to match.  This example is based on the operation $>= 0.5$ (attribute value), with buffer $b$.}
\label{figsoftmargin}
\end{center}
\end{figure}

Parameters of an atom feature using a tanh activation function are initialized as follows:

\begin{align}
c\; &= \frac{b}{1 - \tanh^{-1}{(t)}} \\
w_1 &= \quad\frac{1}{c}\\
w_0 &= - \frac{v}{c} + 1
\end{align}
where $t$ is the activation threshold that defines the feature as matching, $b$ is the threshold buffer, $v$ is the boundary attribute value, $w_1$ is the weight value responding to the input attribute and $w_0$ is the bias value of the unit.  To represent the function `$\geq 0.5$' with a margin $0.1$, set $v = 0.5$ and $b = 0.1$.  This is shown in Figure~\ref{figsoftmargin}.  
Negative ranges are captured as:

\begin{align}
w_1 &= - \frac{1}{c}\\
w_0 &= \frac{v}{c} + 1
\end{align}

Responses for nominal attributes are defined as binary matches, such that the atom responds as matching if a specific attribute value is observed.

The activation function of composite features is defined as a soft-and, based on child activations, as follows:

\begin{align}
b_{comp} &= \sum_n a_n - nt\\
c &= \frac{b_{comp}}{1 - \tanh^{-1}{(t)}} \\
w_{1 \ldots n} &= \quad\frac{1}{c}\\
w_0 &= - \frac{\sum_n a_n}{c} + 1
\end{align}
where $t$ is the activation threshold, $n$ is the number of child elements of this composite and $w_{1 \ldots n}$ are the child weights of the composite.

\subsection{Randomly initialized gradient descent (second method)}

The second method is based on the definition of feature responses using randomized weights.  It does not attempt to capture a relationship with a logical function, and as such allows a greater variety of function definitions, without the ability to maintain a connection with a logical, interpretable function.

Each atom may have a number of child weights, each responding to a given input attribute, which are initialized using a random weight value.  At the time of creation, the weights are adjusted such that the feature provides a positive response for the given input.  This is performed using gradient descent according to the output of the feature, until the activation is above the match threshold.  Composite features are initialized in the same manner.  Ensuring new features provide a positive match value allows the system to operate using the creation and selection methods described previously.

\subsection{Training procedures}
\label{trainingproc}
Several approaches can be used for training.  The feature network can be implemented as either discrete or continuous, training can take place on just the classification layer or with gradient descent throughout the network, and this can operate with or without the population creation and selection mechanism being active.  

The simplest approach is using a discrete feature network, with a dynamic population such that features are added and removed, and performing training on weights between the features and classification nodes.

Using a continuous feature network, the simplest design is to allow continuous activation throughout the network, and perform gradient descent through both the classification layer, through composite features, and on the atomic features.  This approach is slower, as it requires passing activation through the entire network, and subsequently passing gradient descent updates through the network as well.

Alternative training methods can also be used.  Using the soft-logic method, the network can be initially constructed and trained as a discrete network, which can operate faster as selective activation occurs through only active features, and deeper gradient descent is not used.  At a subsequent stage, the network is operated as continuous, using full bottom-up and top-down passes.  For fine-tuning it is possible to disable population dynamics (the coarse learning method), and only operate gradient descent on the fixed topology.

Using the randomly initialized network, it is also possible to use a coarse learning process, and performing initial training only on active features (training only the classification layer, and a dynamic population), before using a full bottom-up and top-down pass, with or without the dynamic population to alter the topology.  It is not recommended to use full gradient descent with only selective bottom-up activation, as the gradient descent will only act to decrease feature activity.

Acting as a discrete network, and only exploring active features, can significantly save time per training instance, however leads to a different exploration of solutions than when using a full gradient descent approach.  The appropriate method to use will depend on the problem being addressed.




\subsection{Cognitive connections}
The design of ADN is largely based on properties seen in studies of cognition, with a number of refinements for practical reasons that diverge from comparable cognitive models.  \citet{akthesis} provides a more detailed description of the cognitive processes used as a motivation for this design, describing a broader model that includes a number of bottom-up, top-down and associative interactions.

Logistic (/logit) and softmax regression are established models of decision making in human cognition, based on examining the decisions made by people based on observed features~\citep{hensher2005applied, daw2006cortical}.  Other models of decision making describe the use of sparse selection of features to form decisions, using an operation more closely related to Lasso regression~\citep{gabaix}.

The original motivation for the method of maintaining a population of features in ADN was based on the reinforcement of memory traces, that are reinforced through use, and decay with time or competition, as described in models such as ACT-R~\citep{anderson04}.  The method described in this paper, based on the magnitude of weights of parent features developed through regression, places more emphasis on the role of features in influencing decisions than on reinforcement through use.  This feature selection approach, and the method of selecting sparse weights, has been chosen for pragmatic reasons, and can be considered more closely related to models of selective feature influence such as~\citep{gabaix} than reinforcement-based models. 


New features in our system are developed based on observed co-occurrences of existing features, an important factor in associative relationships~\citep{mcnamara2005semantic}, and Hebbian learning at the local scale~\citep{oreilly98}. 
These are typically described as direct connections between concepts or stimuli, and examined in terms of priming effects between the stimuli, however the use of association between terms to define a relationship, that is captured as a separate concept, is consistent with these processes.  

The LCS-based design uses a set of independent overlapping classifiers, where a population contains general representations as well as specializations that provide refinements for particular examples.  This is related to cognitive basic level and subordinate categorization, such as a `bird' and a `penguin', where a specialized category captures specific properties that are different from the general class~\citep{tanaka2001entry}, and recognizing general or special classes occurs fairly independently~\citep{scott2006reevaluation}. 

The structure of the feature network shares some common properties with visual and semantic models.  
The use of incrementally selective features capturing more complex structure is related to networks of simple cells, and if \OR operations are introduced there are similarities with the stages of specialization and generalization seen in simple/complex cell networks~\citep{riesenhuber2002neural}.  Re-use of shared features is also seen in models of semantic cognition~\citep{hutchison2003semantic}, which provide further details of associative relationships between concepts and features.  These associative properties are useful for identifying contextual influences, although such priming effects are not included in the presented design.


The amalgamation of these properties does not represent an integrated cognitive `model', however identifying connections between the way the learner operates and behaviours that can be observed in cognitive studies allow information to be exchanged.  
For example when faced with questions about how a particular process should operate, it is possible to examine what is known about the physical property and what clues are available about how it operates, and to capture those properties in the artificial learner.  As a corollary, questions may be raised about how a process should take place, such as whether consistent, moderate activation of all the parts of an object is more or less important than strong activation of some parts with weak (or no) activation of others.  If these questions are not satisfied by existing cognitive studies, they can also provide motivation for experiments that would be useful to conduct.

Capturing properties of these processes and maintaining connections is one of the motivations of the design of the ADN learner, following the integrated design described in~\citep{akthesis}.


\section{UCI and Protein Structure Prediction datasets}
\label{pointfeatures}
In the first set of experiments, we examine the behaviour of the ADN system on a variety of datasets from the UCI collection~\citep{Bache+Lichman:2013}, and Protein Structure Prediction~\citep{stout2008prediction}.  
These datasets use a mix of nominal and real valued attributes, and are the medium and large (and some small) datasets examined previously in~\citep{franco2013gassist}.  
Table~\ref{tabuciproperties} describes the properties of each dataset.  

\begin{table}[h]
\centering
\begin{tabular}{|cccc|}
\hline
\textit{Name} & \textit{Instances} & \textit{Attributes} & \textit{Classes} \\
\hline
pen           & 9,892              & 16 (0,16)           & 10               \\
sat           & 5,792              & 36 (0,36)           & 6                \\
wav           & 4,539              & 40 (0,40)           & 3                \\
SS            & 75,583             & 300 (0,300)         & 3                \\
adult         & 43,960             & 14 (8,6)            & 2                \\
c-4           & 60,803             & 42 (42,0)           & 3                \\
pmx           & 235,929            & 18 (18,0)           & 2                \\
kdd        & 444,619            & 41 (15,26)          & 23               \\
SA            & 493,788            & 270 (26,244)        & 2                \\
CN            & 234,638            & 180 (0,180)         & 2               \\
\hline
\end{tabular}
\caption{Properties of datasets examined.  Attributes describe the total number, followed by the number of discrete (nominal) and real-valued attributes.}
\label{tabuciproperties}
\end{table}

GAssist~\citep{bacardit2004pittsburgh} and BioHEL~\citep{bacardit2009mixed} are Genetics-based Machine Learning (GBML) systems that have been successful for addressing protein-structure prediction and a range of other domains.  GAssist is a form of Learning Classifier System, using the Pittsburgh approach where the Genetic Algorithm operates over a full set of classifiers, rather than acting on individual classifiers as used in Michigan systems~\citep{holland2000learning}.  BioHEL uses Iterative Rule Learning (IRL), another form of GBML.  Both systems use specialized operations such as smart initialization and efficiency enhancement, 
and as such use a more complex design than the base LCS or IRL systems, however the specialization is not considered domain specific.  These refinements allow the systems to be scalable and operate on more complex problems than the base designs.

\citet{franco2013gassist} examined these systems on the following UCI and PSP datasets (and others), along with comparisons with state of the art systems such as Support Vector Machines.  

Testing of our system has been performed using 10-fold leave-one-out cross-validation.  Minimal parameter exploration per problem has been used, on some problems the number of features used has been varied, for example decreasing the amount when over-fitting is seen.  A scheduled learning rate has been used, starting with an initial rate of 0.1 and decreasing as a multiple of 0.95-0.98 each epoch.  

The kdd problem uses a large number of classes (23), which leads to slow performance due to the number of connections between features and class units.  For efficiency a sparse set of connections between features and classes has been used, limiting the number to 1000, described further in Section~\ref{secsparse}. Results are presented in Table~\ref{tabuciresults}.

\begin{table}[h]
\centering
\begin{tabular}{|l|llllllll|lll|}
\hline
\textit{} & \textit{C45} & \textit{NB} & \textit{SVM} & \textit{SVMr}  & \textit{PART}  & \textit{IB5} & \textit{GA} & \textit{Bio} & \textit{Discr} & \textit{GD-R} & \textit{GD-S} \\
\hline
adult     & 85.99        & 83.24       & 84.93        & 83.66          & 85.64          & 82.62        & 86.08       & 86.09        & \textbf{86.26}    & 81.05           & 83.91            \\
c-4 & 80.89        & 72.11       & 75.85        & \textbf{83.95} & 79.22          & 81.03        & 79.77       & 80.94        & 78.93             & 61.25           & 73.09            \\
kdd    & 99.96        & 92.27       & 99.93        & 99.93          & \textbf{99.97} & 99.93        & 99.25       & 99.95        & 99.93             &   99.61              &    99.70          \\
pen       & 96.62        & 85.76       & 97.94        & \textbf{99.45} & 96.89          & 99.22        & 87.27       & 94.94        & 98.51             & 97.60           & 99.07            \\
sat       & 86.68        & 79.63       & 86.88        & 89.35          & 86.54          & 90.91        & 83.74       & 88.14        & 89.70             & 88.60           & \textbf{91.08}   \\
wav       & 74.96        & 79.86       & 86.7         & \textbf{86.7}  & 77.96          & 79.8         & 83.14       & 84.98        & 83.96             & 86.54           & 81.90            \\
PMX       & 76.99        & 48.91       & 49.59        & 76.27          & 59.89          & 87.58        & 87.5        & \textbf{100} & 93.49             & 50           & 50            \\
CN        & 73.14        & 75.68       & 80.18        & \textbf{83.34} & 74.79          & 78.31        & 77.7        & 80.59        & 80.21             & 80.49           & 80.58            \\
SS1       & 55.12        & 67.73       & 73.2         & 71.75          & 58.92          & 58.25        & 62.98       & 71.19        & 71.45             & 74.04           & \textbf{74.47}   \\
SA6       & 70.96        & 75.12       &              &                & 71.88          & 75.53        & 76.76      & 79.3         & 78.74             & 78.96           & \textbf{79.62}  \\
\hline
\end{tabular}
\caption{Table of results of comparison systems with the discrete ADN system, and the randomly-initialized (GD-R) and soft-logic (GD-S) gradient descent ADN systems.}
\label{tabuciresults}
\end{table}

The BioHEL and GAssist methods are specialized systems that incorporate optimizations, as a modification from the base learning algorithm used, allowing them to efficiently address large and complex datasets.  
A further advantage of these systems is that they provide interpretable results, where the classification rules are presented in a readable format, allowing the decision process used to be understood, in contrast to black-box methods such as SVMs and ANNs.
The BioHEL and GAssist systems operate using an ensemble for each test, while for each cross-validation test our system is examined as an isolated learner.  

The results produced by ADN are close to the best results on all datasets, and on a number of measures produce the highest accuracy.  The ADN system is presented in a minimal form, based on a simple algorithm design without specialized refinements, and without the use of ensembles.  As such the ability to produce comparable results with these state of the art systems, and provide consistent high quality results on a wide range of datasets, is significant.

The gradient descent based methods, using soft-logic and random initialization, performed notably better than the discrete method on the SS1, sat, wav and SA6 problems, and showed the highest or near-highest accuracy overall on these problems.  In contrast, on the c-4 and adult problems the discrete method was better, and on the parity-multiplexer problem (pmx) the gradient descent methods failed to learn at all.  This class of problem is examined further in Section~\ref{irregproblem}.  On this problem the gradient descent adjustments interfere with learning, and the population-based methods are more successful.  It is possible to employ a sequential learning approach, using discrete learning prior to gradient-based fine-tuning (see Section~\ref{trainingproc}), however no advantage is expected with gradient learning on this problem.  

The differences in learning behaviour indicate that gradient-based learning only provides advantages on some domains.  In general the Soft-logic and Random initialization methods perform well regardless, and implement a coarse population-based learning method alongside the gradient descent learning, however on specific problems (pmx) the gradient can interfere with learning.

Table~\ref{timing} presents the run-time for reaching the above degree of accuracy on a number of problems, in comparison with the run-time of the systems examined in~\citep{franco2013gassist}.  This has been conducted single-threaded on a 2009 model W3550 Bloomfield Xeon processor, roughly comparable with previous results.  In general runs have been conducted to produce good accuracy within an acceptable time-frame, rather than to emphasize short run-time.

\begin{table}[h]
\begin{tabular}{|l|cccccccccc|}
\hline
\textit{} & \textit{SVM} & \textit{SVMr} & \textit{IB5} & \textit{C4.5} & \textit{NB} & \textit{PART} & \textit{Bio} & \textit{GA} & \textit{Discr} & \textit{GD-R} \\
\hline
adult & 2988 & 20144 & 450 & 14 & 2 & 248 & 1998 & 7875 & 34887 & 131254 \\
c4 & 18219 & 39211 & 1097 & 9 & 2 & 962 & 7197 & 44531 & 63635 & 451792 \\
SS1 & 57133 & 134243 & 26408 & 715 & 168 & 99651 & 33431 & 289783 & 36333 & 71753 \\
kdd & 1347 & 32161 & 96826 & 262 & 168 & 295 & 65807 & 168482 & 48332 & 135454   \\
SA6 &  &  & 146578 & 6073 & 437 &  & 468821  & 271640  & 52311 & 107662 \\
\hline      
\end{tabular}
\caption{Timing of the various systems on the different datasets.  The significant variation in relative timing of the ADN system is due to the choice of schedule, relative performance with time is shown in Figure~\ref{figtime}.}
\label{timing}
\end{table}

\begin{figure}[ht]
\begin{center}
\centerline{\includegraphics[width=.7\columnwidth]{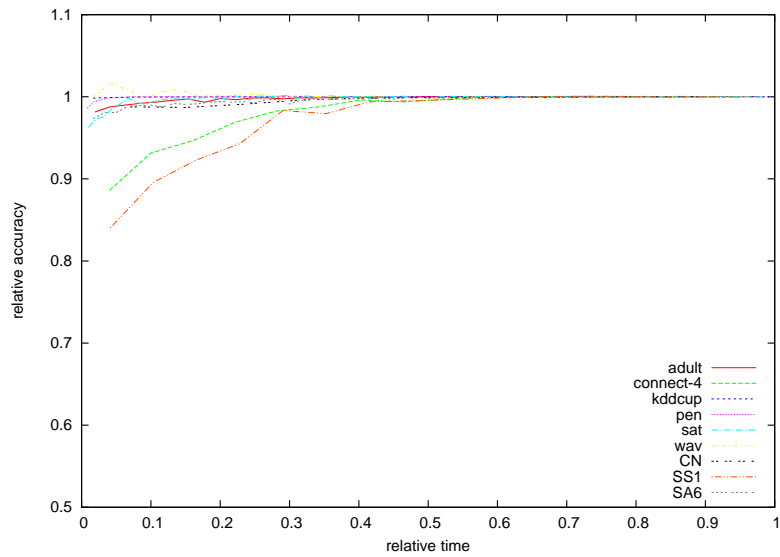}}
\caption{Relative test set performance over time (averaged), according to relative duration of training run.}
\label{figtime}
\end{center}
\end{figure}

\begin{figure}[ht]
\begin{center}
\centerline{\includegraphics[width=.7\columnwidth]{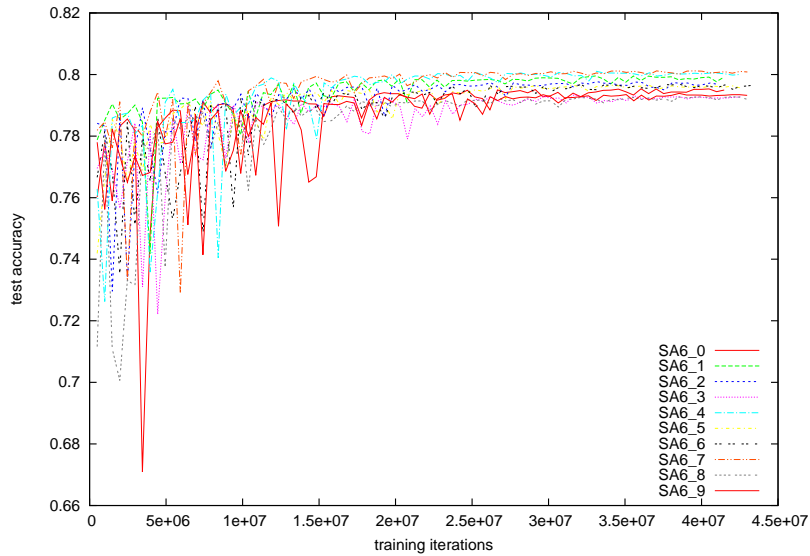}}
\caption{Variation in test accuracy with training, for 10 cross-validation folds of the GD-R learner on the SA6 data set.  The decreasing learning rate schedule allows stability.}
\label{figvariability}
\end{center}
\end{figure} 

These results show some degree of variation, the ADN system took longer on some problems and was faster on others, but overall 
was able to scale well to the larger problems.  
One reason for the variation is based on choices of the amount of time to spend training, which was set a priori, and a long time frame was often used with a diminishing learning rate, to remove variation in test set accuracy for each run (an alternative is to use an ensemble of much shorter runs).  The learning profile is asymptotic and the accuracy level reached is dependent on the amount of time per run, a level with slightly decreased overall accuracy was able to be reached with significantly less time.  The average relative performance with time is shown in Figure~\ref{figtime}, indicating a good solution was found very quickly, and further time was used to give a slight increase in the accuracy.  The test set performance showed a fair amount of variation per epoch in early training in many cases (Figure~\ref{figvariability}), 
in the relative performance chart this variation has been cancelled out by averaging.
The datasets that showed the better timing results were the protein-structure related problems, likely as less variation in test set performance was seen and a shorter schedule could be used.  The pmx problem required the longest time, taking several days to find a solution, possibly due to an inefficient representation, and limitations of the simple random method used for constructing new features.  The gradient-descent methods required more time than the discrete method.  Full gradient-descent has been used each instance, and it is possible to speed-up training by running as the discrete method prior to find-tuning, as described in Section~\ref{trainingproc}.

The ADN system is examined in its basic form, without the use of optimizations used in more specialized systems, and without the use of ensemble techniques.  This demonstrates good scalability of the base learning technique to address large problems.


\begin{figure}[ht]
\begin{center}
\centerline{\includegraphics[width=.7\columnwidth]{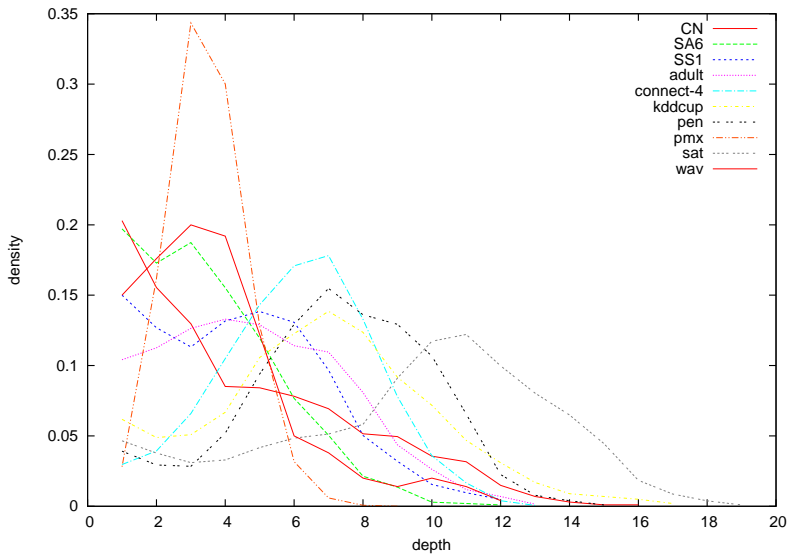}}
\caption{Distribution of the depth of the topology developed, for composite features in each run.}
\label{figucidepths}
\end{center}
\end{figure} 

The distribution of the depth of the feature network developed for each problem is shown in Figure~\ref{figucidepths}.  On problems such as SA6 and CN the feature network has been developed with a shallow topology, indicating selective preference for features with few constraining attributes.  As most features have depth 3 or less, this indicates preference for features with 8 or fewer attributes\footnote{A composite feature of depth 3 can be based on at most 8 atomic features (for example attribute 10 is $\geq$ 2.5), however the number may be less if repeats are used or higher level features link directly to low or atomic features}.  
On the pmx problem most features are at depth 3 or higher, which is reflective of the problem.  The pmx problem combines a 3-bit parity operation with a 6-bit multiplexer, as described in Section~\ref{irregproblem}, which requires a minimum depth of about 3 (using 2 children at each level) to allow discrimination.  On the pen and sat problems the system produced a feature topology with average depth of about 8 and 11, indicating more constrained and specialized features, captured in a larger and more complex network, were useful for finding a solution.  This demonstrates the flexibility of the system to discover an appropriate arrangement of features for the domain being explored.

Common parameters used in these experiments are shown in Table~\ref{uciparams}.

\begin{table}[h]
\begin{tabular}{| p{4cm} p{2cm} p{10cm} |}
\hline
\textit{Name} & \textit{Common value} & \textit{Description}                                                                                                                                                                \\
\hline
\textit{Main hyperparameters} & & \\
Atomic features                      & 500                   & Max size of atomic feature population                                                                                                                                           \\
Composite features                   & 1000                  & Max size of composite feature population                                                                                                                                        \\
Learning rate                        & 0.01                  & Rate of update for logistic/softmax regression, and gradient descent.  In the above experiments an exponentially decreasing rate is used.                                            \\
Creation probability                 & 0.1                   & Probability per instance of creating a new feature (times 0.75/0.25 for incorrect/correct instances).  A new atomic feature is also created based on a further 0.1 probability.                   \\
Removal instances                    & 100                   & Number of instances before trimming population size.  This allows some exploration of new features before removing.                                                                 \\
Maximum children                     & 2                     & Number of child features to use when creating a new composite feature.                                                                                                              \\
                                     &                       &                                                                                                                                                                                     \\
\textit{Optional methods}            &                       &                                                                                                                                                                                     \\
Mini-batch size                      & 1                     & Mini-batch learning is common in neural networks, and acts to average updates over a number of instances.  This is useful on problems such as wav and SA6, however increases learning times and reduces accuracy significantly for others such as binary problems. \\
Sparse out-weight limit           & 0                     & If used, places a limit on the number of weights between features and classifications.                                                                                               \\
Min out-weight depth          & 0                     & It can be useful to restrict feature-classification weights to a minimum depth, if it is known that low level relationships are not useful.                                            \\
                                     &                       &                                                                                                                                                                                     \\
\textit{Gradient descent specific}   &                       &                                                                                                                                                                                     \\
Match threshold                      & 0.5                   & Activation value of units to consider as matching (tanh example).                                                                                                                   \\
Activation function                  & tanh                  &                                                                                                                                                                                     \\
Activation buffer                    & 0.1                   & For soft-logic method (see description).  This value depends on the range of attribute values (with normalizing).     \\                                                         
\hline
\end{tabular}
\caption{Table of common parameters and values used in experiments.}
\label{uciparams}
\end{table}

\subsection{Sparse feature-class connections}
\label{secsparse}
When a large number of output classes are used with a large number of features, the number of feature-class connections can become impractical ($n \times c$).  The following method is used for capturing a sparse relationship between features and classes, by creating and removing feature-class connections in a similar manner to the population methods used for maintaining features.

At each time step a new connection is created between a random active feature and a random class, with a fixed probability or if the number of feature-class weights is less than the limit.  A slight weighting has been introduced, where the weight of choosing a class is proportional to the magnitude of the current delta value for the class (between the current and desired output), to encourage creation of connections for poorly addressed classes.

Using the same interval for removal as the feature population, classification weights are removed with a selection value (`fitness') based on the magnitude of the weight, such that those with low influence are removed.

\section{Irregular binary classification problems and transfer learning}
\label{irregproblem}

A common problem for testing the performance of Learning Classifier and other GBML systems is the binary multiplexer problem, along with a number of related binary datasets.  These problems have been chosen as they have a number of difficult properties, they are multi-modal and epistatic, and gradient descent learning is not effective~\citep{iqbal2012reusing}.  Unsupervised learning cannot be performed as the input is unstructured.  Evolutionary techniques are able to explore the search space in a manner that reliably discovers a good solution, and are able to discover generalizable properties.  The process of recombining elements of partial solutions, with preference for those with high utility, can be effective for exploring this irregular problem space.

\begin{figure}[ht]
\begin{center}
\centerline{\includegraphics[width=0.3\columnwidth]{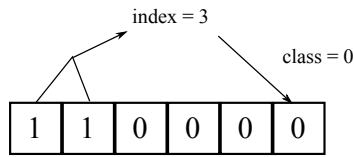}}
\caption{6-bit multiplexer problem.}
\label{figmplx}
\end{center}
\end{figure}

The multiplexer problem is defined using a set of address bits and a set of data bits, where the solution is given as the data bit specified by the address encoding.  For example the six-bit problem uses two address bits and four data bits, as shown in Figure~\ref{figmplx}.  The $n$-bit multiplexer problem is based on $n = k + 2^k$ bits, where $k$ is the number of address bits, and has one bit as the classification class.

The most successful results on this problem have been demonstrated with Michigan style LCS, such as those based on the XCS design~\citep{xcsclassfit}.  \citet{butz2004knowledge} addressed the problem using a form of re-used structure, that develops a decision tree for classification, able to solve the 11-bit problem.  Standard XCS has been shown to solve problems up to size 70-bit.  Specialized designs have solved the 135-bit problem, and are considered to be the state of the art.  The use of a stepped reward function~\citep{butz2005rule} has demonstrated the ability to solve the 135-bit problem, as has XCSCFC~\citep{iqbal2012reusing}, a method based on XCS with re-use of code fragments, without the use of stepped rewards.

Training on the multiplexer problem was conducted with ADN on problem sizes 6, 11, 20, 37, 70 and 135.  
A specific transfer learning operation was introduced, as used with XCSCFC, to allow re-use of features between problems, described further in Section~\ref{transfer}.  This demonstrated solutions on each of the above problems, for the larger problems the results are asymptotic, and as such the solution level has been chosen as 0.995. 

In many LCS systems comparison of results is shown based on the number of instances to reach a solution.  XCSCFC required 500k instances to solve the 70-bit problem, and $2 \times 10^6$ instances to solve the 135-bit problem.  Our ADN system required a larger number of instances to reach a solution, $9 \times 10^6$ instances for the 70-bit problem, and $5 \times 10^7$ instances for the 135-bit problem, however requiring significantly less time per instance, such that the overall time is about the same, of about a week training.  Variations in training instances reflects differences in implementation, partly due to the use of numerous small updates to perform logistic regression learning, on a smaller network of features.  

\subsection{Comparison with XCS}

XCS maintains a population of classifiers using accuracy of prediction for selection (fitness).  ADN is not derived from XCS, however there are design elements that can be compared, as ADN is related to Michigan style classifier systems.  In the following description only the discrete (logical) form of ADN is considered.  

\paragraph{Activation:} When tested against an instance, XCS forms a set of matching classifiers.  In a similar way ADN identifies a set of matching features.  XCS determines an action (classification) based on the weighted average of the active classifiers predicting each class, while ADN determines classification based on a forward pass of the logistic/softmax classification layer.  

\paragraph{Creation:} In XCS new classifiers are created through a covering operation, and through a detailed Genetic Algorithm operation, which selects classifiers according to fitness, constructs children with crossover, and introduces mutations.  Subsumption relationships are tested, by examining if classifiers are generalizations of other classifiers.  In the ADN design, new features are constructed with a fixed probability each time step.  New atoms are constructed based on a random attribute, and new composites are constructed from a random set of $n$ (typically 2) existing features, without weighting.  When adding features, a number of existing features are compared for identity.

\paragraph{Removal:} Removal of classifiers in XCS is based on accuracy of prediction, with further steps taken to preserve the number of classifiers in each niche. In ADN a value is determined for each feature based on the max of the norm of parent weights, and removal of the low valued features occurs after a fixed number of time steps.

Based on the algorithmic description of XCS in~\citep{xcsalg}, in general ADN appears to have lower design complexity than XCS.  

\subsection{Comparison with XCSCFC}

XCSCFC~\citep{iqbal2012reusing} describes another approach for the re-use of elements to define learning rules.  
This approach is described in terms of Genetic Programming, where trees are constructed for classification that combine terminal symbols with logical operators such as (\AND, \OR, `nand' and `nor'), or numeric operators ($+$, $-$, $/$ and $\times$).  A population of such trees are maintained, and recombinations of existing code fragments are used to construct new rules.

XCSCFC uses a population of classifiers, each composed of a fixed number of code fragments, where each code fragment is a binary tree of at most depth two.  The classifier matches if all code fragments match, effectively acting as an \AND operation on the set of code fragments.  As such, a classifier based on six code fragments describes a tree of up to 43 nodes, each node either an operator or terminal symbol.

The operation of XCSCFC is based on XCS, with a number of modifications to the crossover and subsumption procedures, and it does examine classifiers for equality and relative generality.  Generality is not examined in ADN as subsumption procedures are not used.


The code fragments and tree structures in XCSCFC can be considered as related to the feature network of ADN.  In the implementation of ADN used on the 135-bit problem, 10,000 composite features are used, and 270 atomic features, as well as 10,000 classification weights between composite features and the classification unit.  This is effectively a tree of 5,270 nodes.  In XCSCFC 33 code fragments are used for each classifier, and there are 50,000 classifiers in the population.  Although optimizations may be used, the design implies that code fragments are encoded redundantly, in other words a common fragment that is present in multiple classifiers is represented multiple times, and must be activated independently for each classifier.
The density of code fragments is not specified, however the maximum network size of such a population is $((33 \times 7) + 1) \times 50,000$ nodes, and due to the creation process the minimum is presumably $34 \times 50,000$ nodes.  The trees are not directly comparable, however the relative number of nodes used is an indication of the design differences, and the differences in relative time required to examine each instance.  Repetition of code fragments would be present in the GA-based design, while in ADN re-used features occur singularly and are linked with new features to produce new combinations.


\subsection{Transfer learning between problems}
\label{transfer}
One of the advantages of using re-used features, is the ability to discover aspects of one problem, and apply them to a second problem in a manner that improves learning.  On the multiplexer task a specific procedure has been described in~\citep{iqbal2012reusing} to carry features from one problem to another, this operation has also been applied with ADN on the multiplexer problem to allow comparison.  Learning is first conducted on the small 6-bit problem, and after training is finished the most successful classifiers or features in the population are preserved in a set of kept features.  Training is then performed on the next level problem (11-bits), 
features in the kept set are tested for activation each instance, and when creating a new feature, child features are chosen from the kept set with probability 0.5, reflecting a procedure described in~\citep{Banzhaf:1998:GPI:280485}.

While the two systems both address methods for constructing learning rules based on re-used elements, there are differences both in design and the learning approach used.  In XCSCFC, learning is shown to be more successful when a larger number of code fragments per classifier is used.  In general the density of representation is much higher than in that used in the ADN experiments.  A higher number of code fragments, and representation density, implies a more specialized representation per classifier, in other words using a greater number of constraining variables.  Increasing the representation density in ADN did not show improvements, and in general a lower number of constraining variables is used.  This appears to imply a different search strategy, emphasizing different degrees of feature/rule specialization.  

Specialized features in ADN are constructed as combinations of more general features, incrementally adding constraints, however this does not necessarily imply a general to special search strategy must be used.  Specialized features can be developed early on by creating multiple features at a time, for example constructing a random arrangement of features where the highest level features are based on a large number of constraining attributes.  In this way specialized features can be captured early, and more generalized features identified as useful lower level elements that provide similar accuracy, implementing a special to general search strategy.

There may be advantages in using a more complex representation, such as the use of additional Genetic Programming operators and more varied methods for creation of new rules as used in XCSCFC, however the simpler representation has been chosen for consistency with the approach used on the other problems.  A large number of training instances was needed to find an accurate result on the larger problems.  This is a result of the simple creation method used, where new features are constructed from random combinations of existing active features.  The presence of composite features, as useful building blocks, helps to improve the probability of creating useful new features, however on problems such as the 135-bit classifier where many irrelevant features are present, the random creation method requires a lot of exploration to find a good solution.


The ADN implementation on the multiplexer problem is based on the same design used in the experiments in Section~\ref{pointfeatures}, with the introduction of the transfer learning approach to allow comparison.  The ability to address the difficult and irregular multiplexer problem is supportive of the general applicability of the design. 


\section{Image classification with gradient descent}
\label{secimages}
The UCI, Protein Structure datasets and binary problems are domains that are successfully addressed by GBML systems.  On other areas such as image classification gradient descent methods are more successful, and these problems are often not well addressed by evolutionary systems.  This area is explored with our ADN system, to examine advantages of including a coarse, population-based search strategy alongside gradient descent learning.

A common benchmark problem is the MNIST dataset of handwritten digits.  This problem has been well addressed by artificial neural network methods, including deep learning methods, such as convolutional approaches~\citep{lecun98}, Restricted Boltzmann Machine-based methods using Dropout~\citep{hinton2012improving}, and other combinations of approaches including the use of DropConnect~\citep{wan2013regularization}.  Approaches using GBML techniques have been used~\citep{lcsmnist}, however they have not captured the same degree of precision seen in gradient descent methods.

\subsection{Convolutional implementation}

In the design presented in Section~\ref{gradientdescent}, features are considered to either match or not, and in the previously described examples atomic features are bound to a single input attribute.  Some of the most successful artificial neural network implementations are based on a convolutional design, where a feature can match in multiple positions, described as using shared weights to provide responses in different locations.

In existing Convolutional Neural Network (CNN) designs, the topology of the network is pre-defined, and studies of vision-specific CNNs have shown that the choice of topology is a critical aspect of performance.  The use of absolute value rectification, local contrast normalization and average pooling is significant for allowing successful operation~\citep{jarrett2009best}, and such a structure allows top-level results regardless of the method used for developing features.  This occurs to the extent that the use of random features within the given topological structure provides high level results comparable to those with more sophisticated features.  

In this experiment we will examine the use of our Abstract Deep Network system for developing a network topology in a self-organizing manner, using a convolutional structure.  The network is developed using the coarse population-based method to construct the feature network, while gradient descent allows a finer level of learning than that based on the population method alone.  Specific functions such as local contrast normalization and average pooling are not used.

\subsection{Network structure}

\begin{figure}[ht]
\begin{center}
\centerline{\includegraphics[width=\columnwidth]{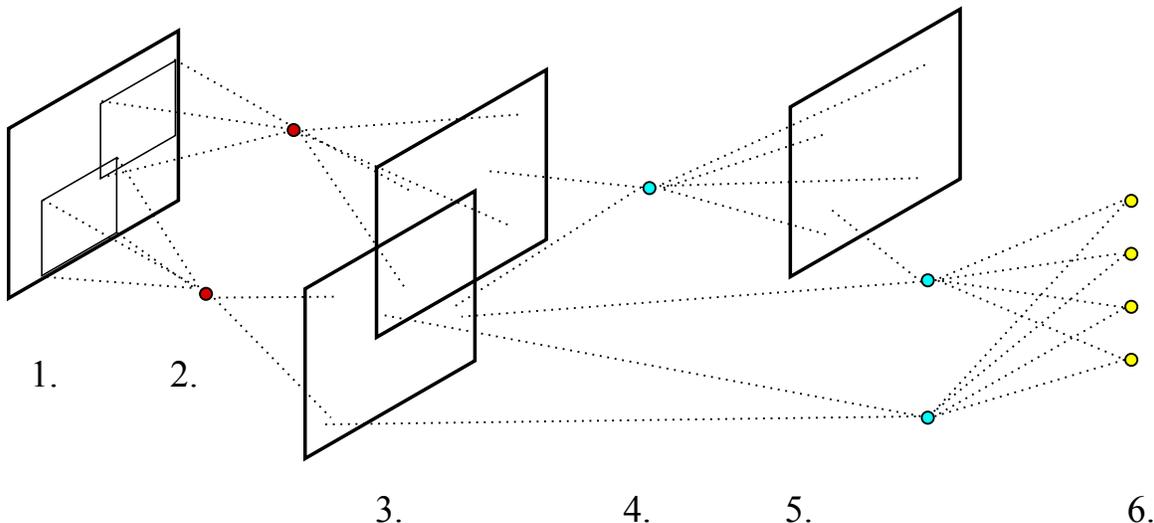}}
\caption{Structure of the feature network, including convolutional maps.  1. observed input, 2. atomic features, 3. activation map, 4. composite features, which are constructed from a number of child atomic or composite features, and respond according to a single point on each child feature map, with relative position defined as a vector, 5. composite activation map, 6. softmax output, connected to all composite features, responding to the max value of each composite activation map. }
\label{figconvnet}
\end{center}
\end{figure}

The key difference between the convolutional and basic design of ADN, is that each feature can be tested for a match in multiple positions, and as such activation values are presented as a map rather than a single activation value.  An assumption of the dimensionality of input is used, and each atomic feature is constructed to respond to a rectangular region rather than a single attribute.  The structure of the network is given in Figure~\ref{figconvnet}.

Weights of new atomic features are created according to an observed region, and are chosen such that the response of the feature to the observation matches a constant value $k$, by initializing weights according to observed values, and normalizing such that the sum of squares is $k$.  Feature activation values are continuous, and in order to capture features being active or inactive, a threshold value is employed.  Identifying matching or non-matching features is used in selecting features for constructing new composite features, while all features are included for passing activations between features and classification units, as described in Section~\ref{gradientdescent}.

When a new composite is created, each child weight is set to the current activation value of the child feature, and the weights normalized such that the square sum is a constant $k$.  By initializing the bias value to $-(k-\epsilon)$, the composite is initialized to be active approximately when the conjunction of the child elements are active.  The selection (`fitness') value used is based on reinforcement through use, based on the average of a reinforcement value $r_t$ at each time step, applied to the feature with the highest predictive value and each its child elements, according to $\Delta f_t = \alpha (r_t - f_{t-1})$.
\footnote{These feature initialization and selection methods are a variation from the methods described previously, the method in Section~\ref{gradientdescent} is preferred.}.

The relationships between features and classifications are captured in fully-connected weights between composites and a `softmax' classification layer.  This produces a topology that is not strictly layered, and is necessary due to the self-organizing topology of composite features. 

As each feature produces an activation value for each position, the mapping between the composite feature layer and the classification layer is based on a single activation value for each feature, using the maximum value.  To perform backpropagation, the position with the highest value in the top-level feature is identified, and the activation value for each child feature relative to this position is used.

\subsection{Convolutional experiments}

The MNIST dataset is based on a set of 70,000 images for training and testing, of size 28x28.  
Training of our system was performed in two phases, first using the population reinforcement method for creation and selection of features, to develop the network topology, at the same time as applying gradient descent learning.  This is the same approach as that used by the gradient descent methods in Section~\ref{pointfeatures}.  This phase was conducted for $2 \times 10^5$ instances, to construct the network topology based on the coarse learning technique.  Subsequent fine-tuning was conducted with the topology fixed, using only gradient descent to adjust the weights.  

Two network structures have been examined.  The first uses a large set of 1500 atomic (first level) features and 5000 composites (approx $2\times 10^5$ weights total), the second uses 100 atomic features and 10,000 composites ($1.4 \times 10^5$ weights).  These configurations are referred to as 1500A-5000C and 100A-10000C respectively.  The first network allows a larger number of low level feature maps, while the second relies on the re-use of a limited set of basic features through composition relationships.

The network with a large set of atomic features showed an average error rate of 0.79\% (classification errors on each run out of 10,000: 63, 69, 71, 71, 80, 81, 81, 83, 83, 83, 89, 93), while the small set network showed an average error of 0.86\%. A randomly generated topology was also tested for comparison, where the topology was constructed in a similar manner, without using active features for selection, and without removal based on the population selection metric.  

The use of a randomly initialized topology showed significant variation in performance, with many runs failing to find a good solution.  Of those that were successful, an average error rate of 1.25\% was shown.  This suggests that in many cases the randomly developed network fails to build a topology that is suitable for allowing subsequent gradient descent learning.  The final result for the randomly constructed network is not as accurate as those constructed using the coarse learning method.  

Results shown are based on Rectified Linear Units (ReLU)~\citep{krizhevsky2012imagenet}, which provide slightly faster convergence and less processing than logistic units, and introduce a rectification process, a factor which was shown to be important for learning on image domains~\citep{jarrett2009best}.  No pre-processing of data has been used, other than to scale inputs with mean zero and $\sigma=1$.

Previous results using standard non-convolutional neural network techniques are typically limited to approximately 1.6\% error, while experiments using Restricted Boltzmann Machine based systems have shown an error rate of 0.95\%, and 0.79\% using Dropout~\citep{hinton2012improving}, described as a record for systems without prior knowledge or enhanced training sets.  Lower error rates have been shown with systems that use significant pre-processing, or are based on a specific topology, for example 0.6\% has been shown using pre-training and sparse feature selection in a convolutional network~\citep{ranzatosparse}, and results as low as 0.21\% have been shown using a mix of convolutional methods with the DropConnect operation~\citep{wan2013regularization}.

The error level shown by the ADN system is similar to that shown by RBM techniques using Dropout, also without the use of enhanced training sets.  The results are not directly comparable, as further assumptions have been incorporated into our design, including an assumption of the dimensionality of the input, and the use of convolution.  Our system does however introduce less assumptions about the domain than existing Convolutional Neural Networks.  The Convolutional Deep Belief Network~\citep{cdbn} provides another approach (0.82\% error), where pre-trained features are used in a convolutional architecture, however this approach focuses on the development of low level features that are used by a Support Vector Machine, with a kernel function specialized towards image domains~\citep{grauman2005pyramid}.  

The topology used by ADN is self-organizing, and the results shown have not used specific functions common to CNNs, such as pooling and local contrast normalization, which are an important aspect of their success~\citep{jarrett2009best}.  Our approach addresses the development of higher level relationships, with very sparse connections between features, identified from observed relationships.   There may be further advantages from combining the approach with existing low-level feature discovery methods and pooling and normalization functions. 

\begin{figure}[ht]
\begin{center}
\centerline{\includegraphics[width=.7\columnwidth]{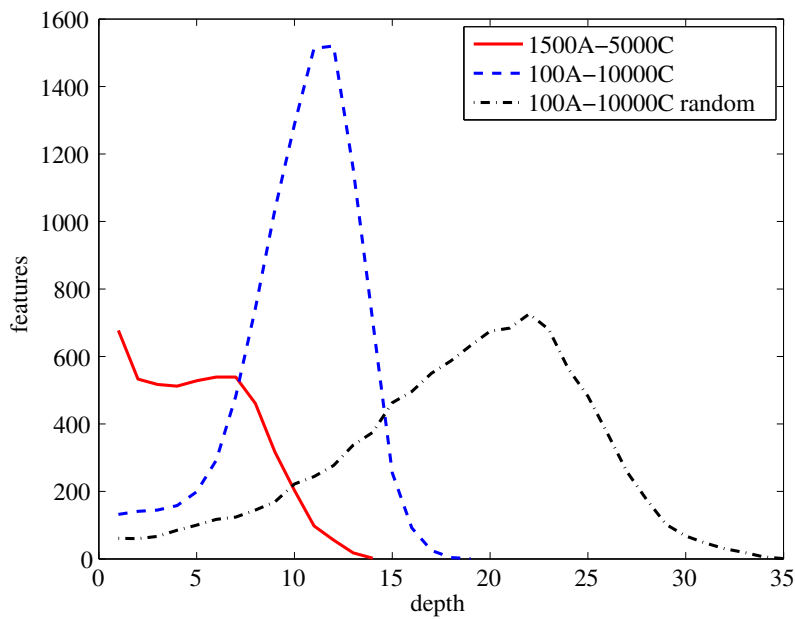}}
\caption{Depth distribution of elements in the developed 1500A-5000C network, and the developed 100A-10000C network.}
\label{netstats}
\end{center}
\end{figure} 

The depth profile of the network produced by ADN for the three network structures is shown in Figure~\ref{netstats}.  This shows that a much flatter network is produced when a larger number of atomic features are available, and greater depth is required to capture features in the 10000C network, requiring further re-use of existing elements.  The topology is significantly different to that used in layer-wise fully connected networks, with a vastly lower number of child connections per feature (except for feature-classification connections), while the depth of the network is greater.

Both of the developed networks show a smaller average depth than the randomly constructed network, 
as a result of a bias for constructing features from co-occurrences of active features, and selective preference towards a shallower network.  This may be indicative of a point where addition of specialization features to existing representations produces structures that are not useful and are not reinforced, related to `terminal features'~\citep{fidler2008similarity}.

Dropout has been shown to improve performance and generality of ANNs at a cost of increased training time, improving the error rate of a standard feedforward network from 1.6\% to about 1.1\%.  Introducing Dropout into our system led to significantly degraded performance.  This may be due to the use of weights initialized to represent conjunction relationships, as inhibited activation of units will lead to greater reduction in activation of parent nodes.  Our network captures sparseness through a different mechanism, and as such the random inhibition behaviour of Dropout does not appear to be beneficial.

This experiment has demonstrated that the coarse learning process based on a population creation and selection mechanism, is able to develop a network topology that allows successful behaviour on a well-studied image classification problem.  Random initialization of the network produced a topology that was unreliable for supporting subsequent gradient descent learning, while the ADN network consistently produced effective learning.  The overall performance demonstrated is in a similar range as established deep neural network techniques, based on an independent learning mechanism.

\section{Conclusions}
\label{secconc}

In this study we have presented the design of a learning technique that 
combines a coarse learning process, related to evolutionary learning, with gradient descent.  The design is motivated by cognitive properties seen in behavioural studies, with the aim of introducing higher level processes into a practical artificial learning technique.  
The cognitive analogy based on behavioural experiments is intended as basis for guiding design, in contrast with the evolutionary paradigm used by GBML systems.  The design is seen as an abstraction of neural-level processes, typically addressed in other artificial neural network systems.

The network is constructed as a population of re-used elements, using observed co-occurrences of features as a basis for the development of new features.  The network topology is constructed in a guided fashion, rather than being defined a priori.  This is important for allowing scalability, as higher level features are constructed using a very sparse set of connections with lower level elements.  In contrast, features in designs such as Convolutional Neural Networks are connected to all features in the next lower layer.

In the current ADN network, higher level elements are specializations of lower level ones, introducing additional constraints, and the semantic basis of elements in the network can be recognized.  An extension of the design that includes generalization relationships (\OR operators) may be beneficial, and allow flexibility, however this has not been currently addressed.  When the network is implemented as a discrete system, using atomic operators such as $\geq$, $\leq$ and $=$, the rules that are developed are interpretable, allowing understanding the solutions developed by the system, in contrast to the black-box approach of many artificial neural network and Support Vector Machine systems.

A diverse range of problem domains have been examined, including medical, signal processing and other datasets from the UCI collection~\citep{Bache+Lichman:2013}, Protein Structure Prediction~\citep{stout2008prediction}, and irregular binary classification problems.  Where possible the system has been tested in a basic form, with minimal optimizations that introduce complexity.  The basic design has demonstrated top-level behaviour compared with refined GBML approaches and standard machine learning systems, and the ability to operate on fairly large datasets in a scalable manner.  When introducing a transfer-learning mechanism similar to that used in~\citep{iqbal2012reusing}, results comparable with the state-of-the-art is shown on a difficult irregular problem, partly due to the ability of the system to operate on large problems efficiently, along with an effective mechanism for preserving and re-using elements of solutions.

The image classification domain is addressed using a convolutional network of features. 
Minimal specializations are introduced, and fewer assumptions are used than existing convolutional networks.  Other systems have shown higher accuracy, however the level reached by the ADN developed network is in a similar range to established systems, and introduces a novel, general, self-organizing method for developing a topology for the image domain based on re-used features.

The feature network can operate in a discrete manner, where representations are described in a logical form and activation values are discrete, or as a continuous system trained using gradient descent.  On some problems such as binary classification problems the discrete system was the most successful, while on others such as protein structure prediction and image classification the gradient descent method was preferred.  A critical view is that the system is simply acting either as a logical Learning Classifier-type System or as an artificial neural network, depending on the problem.  This is not the case, on the image classification problem the use of gradient descent learning on a randomly initialized network was unreliable, and not as accurate as that trained using the coarse, population-based method.  Further, it is possible to allow a gradual transition between a discrete system and a continuous system, by using a separate learning rate for the feature network as for the classification layer, and choosing a value between zero and an appropriate learning rate.  This property is unique, allowing a gradual transition between symbolic and gradient learning, either depending on the problem, or according to different stages of training.


This design aims to connect properties seen in a diverse range of cognitive science areas, and to explore the development of a flexible artificial learning design, combining effects that take place on different scales and levels of abstraction.

In a practical sense, the Abstract Deep Network method allows evolutionary-style learning that is effective for finding a good solution in irregular problem domains, allows the re-use of existing features to assist in development of solutions through recombination of parts, and incorporates fine tuning techniques to find precise solutions based on gradient descent.


\bibliography{features}

\begin{thebibliography}{47}
\providecommand{\natexlab}[1]{#1}
\providecommand{\url}[1]{\texttt{#1}}
\expandafter\ifx\csname urlstyle\endcsname\relax
  \providecommand{\doi}[1]{doi: #1}\else
  \providecommand{\doi}{doi: \begingroup \urlstyle{rm}\Url}\fi

\bibitem[Anderson et~al.(2004)Anderson, Bothell, Byrne, Douglass, Lebiere, and
  Qin]{anderson04}
J.~R. Anderson, D.~Bothell, M.~D. Byrne, S.~Douglass, C.~Lebiere, and Y.~Qin.
\newblock An integrated theory of the mind.
\newblock \emph{Psychological Review}, 111\penalty0 (4):\penalty0 1036--1060,
  2004.
\newblock ISSN 1939-1471.

\bibitem[Bacardit(2004)]{bacardit2004pittsburgh}
Jaume Bacardit.
\newblock \emph{Pittsburgh genetics-based machine learning in the data mining
  era: representations, generalization, and run-time}.
\newblock PhD thesis, Ramon Llull University, Barcelona, Catalonia, Spain,
  2004.

\bibitem[Bacardit and Krasnogor(2009)]{bacardit2009mixed}
Jaume Bacardit and Natalio Krasnogor.
\newblock A mixed discrete-continuous attribute list representation for large
  scale classification domains.
\newblock In \emph{Proceedings of the 11th Annual conference on Genetic and
  evolutionary computation}, pages 1155--1162. ACM, 2009.

\bibitem[Bacardit et~al.(2009)Bacardit, Burke, and Krasnogor]{Bacardit2009}
Jaume Bacardit, Edmund~K. Burke, and Natalio Krasnogor.
\newblock Improving the scalability of rule-based evolutionary learning.
\newblock \emph{Memetic Computing}, 1\penalty0 (1):\penalty0 55--67, March
  2009.
\newblock \doi{10.1007/s12293-008-0005-4}.

\bibitem[Bache and Lichman(2013)]{Bache+Lichman:2013}
K.~Bache and M.~Lichman.
\newblock {UCI} machine learning repository, 2013.
\newblock URL \url{http://archive.ics.uci.edu/ml}.

\bibitem[Banzhaf et~al.(1998)Banzhaf, Francone, Keller, and
  Nordin]{Banzhaf:1998:GPI:280485}
Wolfgang Banzhaf, Frank~D. Francone, Robert~E. Keller, and Peter Nordin.
\newblock \emph{Genetic Programming: An Introduction: on the Automatic
  Evolution of Computer Programs and Its Applications}.
\newblock Morgan Kaufmann Publishers Inc., San Francisco, CA, USA, 1998.
\newblock ISBN 1-55860-510-X.

\bibitem[Bar(2004)]{bar04}
M.~Bar.
\newblock Visual objects in context.
\newblock \emph{Nature Reviews Neuroscience}, 5\penalty0 (8):\penalty0
  617--629, 2004.
\newblock ISSN 1471-003X.

\bibitem[Bengio(2009)]{bengio2009learning}
Yoshua Bengio.
\newblock Learning deep architectures for {AI}.
\newblock \emph{Foundations and trends in Machine Learning}, 2\penalty0
  (1):\penalty0 1--127, 2009.

\bibitem[Butz and Wilson(2001)]{xcsalg}
Martin Butz and Stewart~W. Wilson.
\newblock An algorithmic description of {XCS}.
\newblock In \emph{Revised Papers from the Third International Workshop on
  Advances in Learning Classifier Systems}, pages 253--272, London, UK, 2001.
  Springer-Verlag.
\newblock ISBN 3-540-42437-7.

\bibitem[Butz and Herbort(2008)]{butz2008context}
Martin~V Butz and Oliver Herbort.
\newblock Context-dependent predictions and cognitive arm control with {XCSF}.
\newblock In \emph{Proceedings of the 10th annual conference on Genetic and
  evolutionary computation}, pages 1357--1364. ACM, 2008.

\bibitem[Butz(2005)]{butz2005rule}
M.V. Butz.
\newblock \emph{Rule-Based Evolutionary Online Learning Systems: A Principled
  Approach to LCS Analysis and Design}.
\newblock Studies in Fuzziness and Soft Computing. Springer, 2005.
\newblock ISBN 9783540253792.

\bibitem[Butz et~al.(2004)Butz, Lanzi, Llor{\`a}, and
  Goldberg]{butz2004knowledge}
M.V. Butz, P.L. Lanzi, X.~Llor{\`a}, and D.E. Goldberg.
\newblock {Knowledge extraction and problem structure identification in XCS}.
\newblock In \emph{Parallel Problem Solving from Nature}, pages 1051--1060.
  Springer, 2004.

\bibitem[Chase and Simon(1973)]{chase1973perception}
W.~G. Chase and H.~A. Simon.
\newblock Perception in chess.
\newblock \emph{Cognitive psychology}, 4\penalty0 (1):\penalty0 55--81, 1973.

\bibitem[Daw et~al.(2006)Daw, O'Doherty, Dayan, Seymour, and
  Dolan]{daw2006cortical}
Nathaniel~D Daw, John~P O'Doherty, Peter Dayan, Ben Seymour, and Raymond~J
  Dolan.
\newblock Cortical substrates for exploratory decisions in humans.
\newblock \emph{Nature}, 441\penalty0 (7095):\penalty0 876--879, 2006.

\bibitem[Didierjean and Gobet(2008)]{didierjean08}
A.~Didierjean and F.~Gobet.
\newblock {S}herlock {H}olmes - an expert's view of expertise.
\newblock \emph{British Journal of Psychology}, 99:\penalty0 109--125, 2008.

\bibitem[Erhan et~al.(2010)Erhan, Bengio, Courville, Manzagol, Vincent, and
  Bengio]{erhan2010does}
Dumitru Erhan, Yoshua Bengio, Aaron Courville, Pierre-Antoine Manzagol, Pascal
  Vincent, and Samy Bengio.
\newblock Why does unsupervised pre-training help deep learning?
\newblock \emph{The Journal of Machine Learning Research}, 11:\penalty0
  625--660, 2010.

\bibitem[Fidler et~al.(2008)Fidler, Boben, and Leonardis]{fidler2008similarity}
S.~Fidler, M.~Boben, and A.~Leonardis.
\newblock Similarity-based cross-layered hierarchical representation for object
  categorization.
\newblock In \emph{Computer Vision and Pattern Recognition, 2008. CVPR 2008.
  IEEE Conference on}, pages 1--8. IEEE, 2008.

\bibitem[Franco et~al.(2013)Franco, Krasnogor, and Bacardit]{franco2013gassist}
Mar{\'\i}a~A Franco, Natalio Krasnogor, and Jaume Bacardit.
\newblock {GAssist} vs. {BioHEL}: critical assessment of two paradigms of
  genetics-based machine learning.
\newblock \emph{Soft Computing}, 17\penalty0 (6):\penalty0 953--981, 2013.

\bibitem[Gabaix(2014)]{gabaix}
Xavier Gabaix.
\newblock A sparsity-based model of bounded rationality.
\newblock \emph{Quarterly Journal of Economics}, 2014.

\bibitem[Goldberg(2002)]{goldberg}
David~E. Goldberg.
\newblock \emph{The Design of Innovation : Lessons from and for Competent
  Genetic Algorithms}.
\newblock Kluwer, Boston, 2002.

\bibitem[Grauman and Darrell(2005)]{grauman2005pyramid}
Kristen Grauman and Trevor Darrell.
\newblock The pyramid match kernel: Discriminative classification with sets of
  image features.
\newblock In \emph{Computer Vision, 2005. ICCV 2005. Tenth IEEE International
  Conference on}, volume~2, pages 1458--1465. IEEE, 2005.

\bibitem[Haykin(2009)]{haykin2009neural}
S.S. Haykin.
\newblock \emph{{Neural networks and learning machines}}.
\newblock Prentice Hall, 2009.

\bibitem[Hecht-Nielsen(1989)]{hecht1989theory}
Robert Hecht-Nielsen.
\newblock Theory of the backpropagation neural network.
\newblock In \emph{Neural Networks, 1989. IJCNN., International Joint
  Conference on}, pages 593--605. IEEE, 1989.

\bibitem[Hensher et~al.(2005)Hensher, Rose, and Greene]{hensher2005applied}
D.A. Hensher, J.M. Rose, and W.H. Greene.
\newblock \emph{Applied Choice Analysis: A Primer}.
\newblock Applied Choice Analysis: A Primer. Cambridge University Press, 2005.
\newblock ISBN 9780521605779.

\bibitem[Hinton et~al.(2012)Hinton, Srivastava, Krizhevsky, Sutskever, and
  Salakhutdinov]{hinton2012improving}
Geoffrey~E Hinton, Nitish Srivastava, Alex Krizhevsky, Ilya Sutskever, and
  Ruslan~R Salakhutdinov.
\newblock Improving neural networks by preventing co-adaptation of feature
  detectors.
\newblock \emph{arXiv preprint arXiv:1207.0580}, 2012.

\bibitem[Holland et~al.(2000)Holland, Booker, Colombetti, Dorigo, Goldberg,
  Forrest, Riolo, Smith, Lanzi, Stolzmann, et~al.]{holland2000learning}
John~H Holland, Lashon~B Booker, Marco Colombetti, Marco Dorigo, David~E
  Goldberg, Stephanie Forrest, Rick~L Riolo, Robert~E Smith, Pier~Luca Lanzi,
  Wolfgang Stolzmann, et~al.
\newblock What is a {L}earning {C}lassifier {S}ystem?
\newblock In \emph{Learning Classifier Systems}, pages 3--32. Springer, 2000.

\bibitem[Hubel and Wiesel(1968)]{hubel1968receptive}
David~H Hubel and Torsten~N Wiesel.
\newblock Receptive fields and functional architecture of monkey striate
  cortex.
\newblock \emph{The Journal of physiology}, 195\penalty0 (1):\penalty0
  215--243, 1968.

\bibitem[Hutchison(2003)]{hutchison2003semantic}
K.A. Hutchison.
\newblock {Is semantic priming due to association strength or feature overlap?
  A microanalytic review}.
\newblock \emph{Psychonomic Bulletin \& Review}, 10\penalty0 (4):\penalty0
  785--813, 2003.

\bibitem[Iqbal et~al.(2012)Iqbal, Browne, and Zhang]{iqbal2012reusing}
Muhammad Iqbal, Will~N Browne, and Mengjie Zhang.
\newblock Reusing building blocks of extracted knowledge to solve complex,
  large-scale boolean problems.
\newblock \emph{IEEE Transactions on Evolutionary Computation}, 2012.

\bibitem[Jarrett et~al.(2009)Jarrett, Kavukcuoglu, Ranzato, and
  LeCun]{jarrett2009best}
Kevin Jarrett, Koray Kavukcuoglu, Marc’Aurelio Ranzato, and Yann LeCun.
\newblock What is the best multi-stage architecture for object recognition?
\newblock In \emph{Computer Vision, 2009 IEEE 12th International Conference
  on}, pages 2146--2153. IEEE, 2009.

\bibitem[Knittel(2013)]{akthesis}
Anthony Knittel.
\newblock \emph{Abstract representations in deep networks, capturing rapid and
  top-down cognitive processes in artificial learning}.
\newblock PhD thesis, University of New South Wales, 2013.

\bibitem[Krizhevsky et~al.(2012)Krizhevsky, Sutskever, and
  Hinton]{krizhevsky2012imagenet}
Alex Krizhevsky, Ilya Sutskever, and Geoff Hinton.
\newblock Imagenet classification with deep convolutional neural networks.
\newblock In \emph{Advances in Neural Information Processing Systems 25}, pages
  1106--1114, 2012.

\bibitem[Kukenys et~al.(2011)Kukenys, Browne, and Zhang]{lcsmnist}
Ignas Kukenys, Will Browne, and Mengjie Zhang.
\newblock Transparent, online image pattern classification using a {L}earning
  {C}lassifier {S}ystem.
\newblock In Cecilia Di~Chio, Stefano Cagnoni, Carlos Cotta, Marc Ebner, Anikó
  Ekárt, Anna~I. Esparcia-Alcázar, Juan~J. Merelo, Ferrante Neri, Mike
  Preuss, Hendrik Richter, Julian Togelius, and Georgios~N. Yannakakis,
  editors, \emph{Applications of Evolutionary Computation}, volume 6624 of
  \emph{Lecture Notes in Computer Science}, pages 183--193. Springer Berlin
  Heidelberg, 2011.
\newblock ISBN 978-3-642-20524-8.

\bibitem[Le{C}un et~al.(1998)Le{C}un, Bottou, Bengio, and Haffner]{lecun98}
Y.~Le{C}un, L.~Bottou, Y.~Bengio, and P.~Haffner.
\newblock Gradient-based learning applied to document recognition.
\newblock \emph{Proceedings of the IEEE}, 86\penalty0 (11):\penalty0
  2278--2324, nov 1998.

\bibitem[Lee et~al.(2009)Lee, Grosse, Ranganath, and Ng]{cdbn}
Honglak Lee, Roger Grosse, Rajesh Ranganath, and Andrew~Y. Ng.
\newblock Convolutional deep belief networks for scalable unsupervised learning
  of hierarchical representations.
\newblock In \emph{International Conference on Machine Learning ({ICML})},
  pages 609--616, New York, NY, USA, 2009. ACM.
\newblock ISBN 978-1-60558-516-1.

\bibitem[Llor{\`a} et~al.(2007)Llor{\`a}, Reddy, Matesic, and
  Bhargava]{llora2007towards}
Xavier Llor{\`a}, Rohith Reddy, Brian Matesic, and Rohit Bhargava.
\newblock Towards better than human capability in diagnosing prostate cancer
  using infrared spectroscopic imaging.
\newblock In \emph{Proceedings of the 9th annual conference on Genetic and
  evolutionary computation}, pages 2098--2105. ACM, 2007.

\bibitem[McNamara(2005)]{mcnamara2005semantic}
T.P. McNamara.
\newblock \emph{{Semantic priming: Perspectives from memory and word
  recognition}}.
\newblock Psychology Press, 2005.

\bibitem[O'Reilly(1998)]{oreilly98}
Randall~C. O'Reilly.
\newblock Six principles for biologically based computational models of
  cortical cognition.
\newblock \emph{Trends in Cognitive Sciences}, 2\penalty0 (11):\penalty0 455 --
  462, 1998.
\newblock ISSN 1364-6613.
\newblock \doi{DOI: 10.1016/S1364-6613(98)01241-8}.

\bibitem[Ranzato et~al.(2006)Ranzato, Poultney, Chopra, Cun,
  et~al.]{ranzatosparse}
Marc~Aurelio Ranzato, Christopher Poultney, Sumit Chopra, Yann~L Cun, et~al.
\newblock Efficient learning of sparse representations with an energy-based
  model.
\newblock In \emph{Advances in neural information processing systems}, pages
  1137--1144, 2006.

\bibitem[Riesenhuber and Poggio(2002)]{riesenhuber2002neural}
M.~Riesenhuber and T.~Poggio.
\newblock Neural mechanisms of object recognition.
\newblock \emph{Current opinion in neurobiology}, 12\penalty0 (2):\penalty0
  162--168, 2002.

\bibitem[Scott et~al.(2006)Scott, Tanaka, Sheinberg, and
  Curran]{scott2006reevaluation}
L.S. Scott, J.W. Tanaka, D.L. Sheinberg, and T.~Curran.
\newblock A reevaluation of the electrophysiological correlates of expert
  object processing.
\newblock \emph{Journal of cognitive neuroscience}, 18\penalty0 (9):\penalty0
  1453--1465, 2006.

\bibitem[Si and Zhu(2013)]{si2013}
Zhangzhang Si and Song-Chun Zhu.
\newblock Learning {AND-OR} templates for object recognition and detection.
\newblock \emph{IEEE Transactions on Pattern Analysis and Machine
  Intelligence}, 99:\penalty0 1, 2013.

\bibitem[Stanley and Miikkulainen(2002)]{stanley2002evolving}
Kenneth~O Stanley and Risto Miikkulainen.
\newblock Evolving neural networks through augmenting topologies.
\newblock \emph{Evolutionary computation}, 10\penalty0 (2):\penalty0 99--127,
  2002.

\bibitem[Stout et~al.(2008)Stout, Bacardit, Hirst, and
  Krasnogor]{stout2008prediction}
Michael Stout, Jaume Bacardit, Jonathan~D Hirst, and Natalio Krasnogor.
\newblock Prediction of recursive convex hull class assignments for protein
  residues.
\newblock \emph{Bioinformatics}, 24\penalty0 (7):\penalty0 916--923, 2008.

\bibitem[Tanaka(2001)]{tanaka2001entry}
J.W. Tanaka.
\newblock The entry point of face recognition: evidence for face expertise.
\newblock \emph{Journal of Experimental Psychology: General}, 130\penalty0
  (3):\penalty0 534--543, 2001.

\bibitem[Wan et~al.(2013)Wan, Zeiler, Zhang, Cun, and
  Fergus]{wan2013regularization}
Li~Wan, Matthew Zeiler, Sixin Zhang, Yann~L Cun, and Rob Fergus.
\newblock Regularization of neural networks using {D}rop{C}onnect.
\newblock In \emph{Proceedings of the 30th International Conference on Machine
  Learning (ICML-13)}, pages 1058--1066, 2013.

\bibitem[Wilson(1995)]{xcsclassfit}
Stewart~W. Wilson.
\newblock Classifier fitness based on accuracy.
\newblock \emph{Evolutionary Computation}, 3\penalty0 (2):\penalty0 149--175,
  1995.
\newblock \doi{10.1162/evco.1995.3.2.149}.

\end{thebibliography}

\end{document}